\documentclass[12pt,final]{l4dc2021} 

\usepackage[utf8]{inputenc} 
\usepackage[T1]{fontenc}    
\usepackage{hyperref}       
\usepackage{url}            
\usepackage{booktabs}       
\usepackage{amsfonts}       
\usepackage{nicefrac}       
\usepackage{microtype}      
\usepackage{graphicx}
\usepackage{xcolor}
\usepackage{amsmath}

\makeatletter
\let\Ginclude@graphics\@org@Ginclude@graphics 
\makeatother

\title[Traffic Forecasting using V2V Communication]{Traffic Forecasting using Vehicle-to-Vehicle Communication}
\usepackage{times}

\author{%
 \Name{Steven Wong} \Email{scw039@ucsd.edu}\\
 \addr Department of Electrical and Computer Engineering, University of California, San Diego, CA 92110, USA
 \AND
 \Name{Lejun Jiang} \Email{lejunj@umich.edu}\\
 \addr Department of Mechanical Engineering, University of Michigan, Ann Arbor, MI 48109, USA\\
 \addr School of Engineering and Applied Science, University of Pennsylvania, Philadelphia, PA 19104, USA
 \AND
 \Name{Robin Walters} \Email{r.walters@northeastern.edu}\\
 \addr Khoury College of Computer Science, Northeastern University, Boston, MA 02115, USA
 \AND
 \Name{Tam{\'{a}}s G. Moln{\'{a}}r} \Email{molnart@umich.edu}\\
 \addr Department of Mechanical Engineering, University of Michigan, Ann Arbor, MI 48109, USA\\
 \addr Department of Mechanical and Civil Engineering, California Institute of Technology, Pasadena, CA 91125, USA
 \AND
 \Name{G{\'{a}}bor Orosz} \Email{orosz@umich.edu}\\
 \addr Department of Mechanical Engineering, University of Michigan, Ann Arbor, MI 48109, USA\\
 \addr Department of Civil and Environmental Engineering, University of Michigan, Ann Arbor, MI 48109,USA
\AND
 \Name{Rose Yu} \Email{roseyu@ucsd.edu}\\
 \addr Department of Computer Science, University of California, San Diego, CA 92110, USA \\
 \addr Khoury College of Computer Science, Northeastern University, Boston, MA 02115, USA
}

\begin{document}

\maketitle

\begin{abstract}%
      We take the first step in using vehicle-to-vehicle (V2V) communication to provide real-time on-board traffic predictions.
    In order to best utilize real-world V2V communication data, we integrate first principle  models with deep learning. Specifically, we train  recurrent neural networks  to improve the predictions given by first principle models.
    Our approach is able to predict the velocity of individual vehicles up to a minute into the future with improved accuracy over first principle-based baselines.
    We conduct a comprehensive study  to evaluate different methods of integrating first principle models with deep learning techniques.  The source code for our models is available at \url{https://github.com/Rose-STL-Lab/V2V-traffic-forecast}.
\end{abstract}


\vspace{5mm}
\begin{figure}[!hb]
    \centering
    \includegraphics[width=135mm]{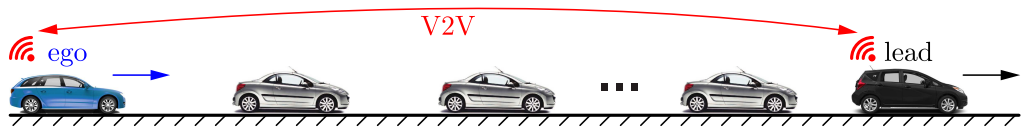}
    \caption{Connected vehicles traveling on the highway where an ego vehicle intends to predict its future motion based on a lead vehicle's past data.}
    \label{fig:V2V}
\end{figure}

\section{Introduction}


The ability to predict future slowdowns on highways in the
timescale of minutes can have significant benefits for traffic participants traveling on the road.
For example, vehicles may use such information to brake earlier and drive more smoothly, improving safety, comfort, fuel economy, and overall traffic throughput \citep{ge2018}. 
Existing traffic forecasting methods mostly rely on collecting data from a large number of vehicles (via loop detectors, cameras, and cell phones), aggregating this data on back-end servers, and using complex models in order to predict the future traffic states. This can capture large scale traffic dynamics but is neither accurate nor fast enough to provide real-time predictions for individual vehicles. In this paper, we consider a novel approach where even single vehicle data can be used to generate predictions in an efficient manner, allowing us to generate predictions on-board in real-time tailored to the needs of individual traffic participants. When implemented on real vehicles,
the method has the potential to  transform the way traffic predictions are generated and utilized.

The main concept is illustrated in \autoref{fig:V2V}, where a connected vehicle, called {\em ego}, obtains information (position and speed data) from another connected vehicle ahead, called {\em lead}.
The lead car's past data may then help to predict the future of the ego car, since the ego will encounter the traffic the lead has already met. 
For example, if the lead car's velocity decreases due to a traffic congestion, the ego car is also likely to slow down when it reaches the congestion wave. 
Such prediction can be done by using first principle-based models that capture the propagation of congestion waves along the highway. 
Alternatively, data-driven methods may be used to obtain predictions, and one may combine them with first principle-based models --- this approach is explored in the rest of this paper.


We use a recently collected  dataset from connected vehicles in real-world traffic \citep{Molnar2021trc} to generate traffic forecasts up to one minute ahead.
Related work makes longer term large-scale traffic predictions using loop detector data \citep{ma2017learning, li2018diffusion}, or short term predictions using camera or Lidar data \citep{wang2020v2vnet, chang2019argoverse}.
To make use of lead car data, it is critical to understand how traffic conditions have evolved since the lead car experienced them. 
While mechanistic models \citep{Bando1995,Treiber2000} are traditionally used to understand the physical principles that govern traffic flow, data-driven methods have also gained popularity recently. 
Purely data-driven methods such as those of \cite{ma2017learning} or \cite{li2018diffusion} use deep neural networks to predict slowdowns, but with limited temporal and spatial resolution.

In this paper,  we propose an integrated approach where recurrent neural networks are trained to learn and correct the errors in the prediction of first principle models.
Unlike a pure deep learning approach, our method leverages first principles from physics and insights from the study of traffic flow.
At the same time, machine learning may discover higher order correlations that are not captured by the first principle model, tune the model to the specific traffic conditions at hand, or improve robustness to unprocessed noisy signals.
Our method achieves better accuracy than either a purely first principle-based baseline or a pure machine learning approach using similar input features.
\paragraph{Contributions}
\begin{itemize}
    \setlength\itemsep{0em}
    \item Utilize a recently collected vehicle-to-vehicle communication dataset to generate high-resolution traffic forecasts.
    \item Combine first principle models and deep learning to achieve better accuracy than either alone.
    \item Investigate the generalizability of the method across different traffic conditions. 
\end{itemize}

\section{Background in Traffic Models}\label{sec:background}

Traffic models have traditionally been
established based on physical first principles.
Although first principle models can capture large-scale traffic dynamics, they often fail to capture the small-scale variability and the uncertainty of human behavior.
More recently, deep learning methods  have also had success in predicting traffic. However, these purely data-driven methods sometimes make unrealistic predictions and have difficulties in generalization.      

\paragraph{First Principle Models}
Describing and predicting the motion of vehicles in traffic has a long history, initially focusing on first principle models.
These include, on one hand, car-following models such as the optimal velocity model~\citep{Bando1995}, intelligent driver model~\citep{Treiber2000}, and models with time delays~\citep{Igarashi2001, Orosz2010}. These are often referred to as microscopic models as they aim to describe the behavior of individual vehicles. On the other hand, traffic flow models such as the LWR model~\citep{Lighthill1955, Richards1956} and its novel formulations~\citep{Laval2013}, the cell transmission model~\citep{Daganzo1994}, and the ARZ model~\citep{Aw2000, Zhang2002} aim for describing the aggregate traffic behavior. These are often called macroscopic models.

In what follows, we will use one of the most elementary car-following models introduced by~\cite{Newell2002}.  Newell's model assumes that each vehicle copies the motion of its predecessor with a shift in time and space that is caused by the propagation of congestion waves in traffic.
This can qualitatively capture an upcoming slowdown, and provide a rudimentary traffic preview for the ego vehicle.
One can potentially improve prediction via more sophisticated first principle models, however, the uncertainty of human driver behavior makes it challenging to achieve low prediction errors.
Alternatively, one may integrate these first principle models, which capture the essential features of traffic dynamics (like the propagation of congestion waves captured by Newell's model), with data-driven approaches in order to improve the quality of predictions.

\paragraph{Deep Learning Models}
Recently,   data-driven approaches such as deep learning have attracted considerable attention for modeling both aggregated (macroscopic) traffic behavior \citep{li2018diffusion,yu2018spatio} and individual (microscopic) driver behavior \citep{wang2018carfollowing, wu2018stabilizing, tang2019multiple, ji2020delayfnn}.  We refer readers to  \citet{veres2019deep} for a comprehensive survey on deep learning for intelligent transportation systems.

Using V2V communication to make traffic predictions with deep learning is a very new area.
\citet{wang2020v2vnet} use V2V communication for perception around obstacles and for making short-term trajectory predictions in urban environments.
\cite{liang2020learning, gao2020vectornet, chang2019argoverse} similarly make short-term predictions on the order of 3 seconds for individual car trajectories in urban traffic using sensor data and not V2V data.    
Meanwhile, our work makes intermediate-term estimates (10-40 seconds) of connected vehicle trajectories on highways.
These approaches consider individualized traffic forecasting for a given vehicle.

Alternatively, many works provide predictions for certain road locations rather than certain vehicles.
\cite{ma2017learning} uses 2D CNNs to make longer-term traffic predictions (10-20 minutes) on various road networks.
\cite{li2018diffusion} models traffic as a diffusion process using recurrent convolutional architecture.
Alternatively, \cite{cui2019traffic} use recurrent graph convolutional networks, \cite{Loumiotis2018}
consider general regression neural networks, and \cite{Yin2018} frame traffic prediction as a classification problem.
\cite{zhao2017lstm} use a spatio-temporal LSTM to make traffic predictions on a similarly long timescale. More recently, attention mechanisms have been shown to be useful for forecasting on this time-scale \citep{do2019effective}.
These models all focus on road networks, so they cannot capture the behavior of individual vehicles in the flow. Their predictions are lower in spatial and temporal resolution than our model, since they rely on lower resolution road sensor data instead of V2V data.
Nevertheless the widespread use of recurrent methods suggests their value in this domain.  


\section{Experimental Data Collected using Connected Vehicles in Highway Traffic}

\begin{figure}
    \centering
     \includegraphics[width=130mm]{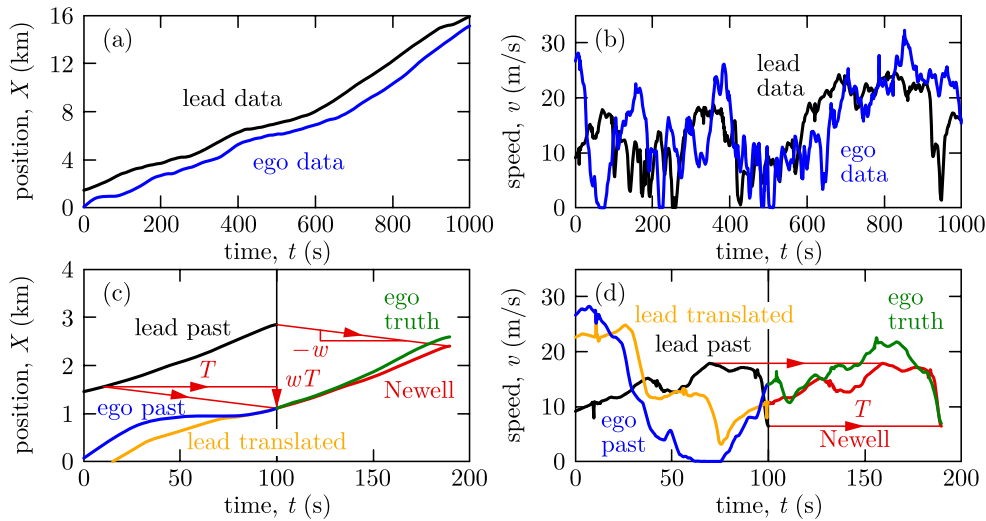}
    \caption{ Experimental position (a) and speed (b) data of two connected vehicles called lead and ego.
    Position (c) and speed (d) predictions from the first-principle Newell's model.}
    \label{fig:shifted_prediction}
\end{figure}

The data we use in this paper were collected by driving five connected vehicles along US39 for three hours in peak-hour traffic near Detroit, Michigan; see~\cite{Molnar2021trc} for details.
Each vehicle measured its position and velocity by GPS, which were sampled and transmitted amongst the vehicles every 0.1 seconds using commercially available devices and standardized broadcast-and-catch protocol.
This dataset is unique as it contains V2V trajectory data of multiple vehicles traveling on the same route for an extended duration of time involving entire traffic jams.
Two vehicles travelled farther ahead of the other three with an average distance of around 1300 and 900 m.
This separation created six potential lead-ego vehicle pairs.
We  consider two vehicle pairs in our study:
the foremost lead vehicle and two different ego vehicles from the group of three.
We omit the data of other vehicles since they were too close to each other to provide long enough predictions.

\autoref{fig:shifted_prediction}(a,b) show the experimental position data (along the highway) and speed data for one of the lead-ego pairs.
The data covers 1000 seconds and includes various traffic conditions.
Since the ego vehicle undergoes qualitatively similar speed fluctuations as the lead, it allows the ego to predict its future motion based on the lead vehicle's past data obtained through connectivity.

\section{Methods}


Given the position data ${X_{\rm L},X_{\rm E} \in \mathbb{R}}$ and speed data ${v_{\rm L}, v_{\rm E} \in \mathbb{R}}$ of the lead and ego vehicles, respectively, we make a prediction at time $t$ about the ego vehicle's speed in the future $\theta$ time ahead.
We denote the prediction by $\hat{v}_{\rm E}(t,\theta)$, that approximates the actual future speed, denoted by $v_{\rm E}(t+\theta)$.

\paragraph{Baselines}
The simplest first principle-based prediction algorithm is the constant speed prediction
\begin{equation}\label{eq:const_pred_cont}
\hat{v}_{\rm E}(t,\theta) = v_{\rm E}(t),
\end{equation}
which does not use data from the lead vehicle, but may work well for short term predictions.

To leverage the data from V2V connectivity, we consider Newell's car-following model \citep{Newell2002} as another baseline.
According to this model, the position $X_{\rm E}$ of the ego vehicle is the same as the position $X_{\rm L}$ of the lead vehicle with a shift $T$ in time and a shift $w T$ in space:
\begin{equation}
X_{\rm E}(t) = X_{\rm L}(t-T(t)) - w T(t),
\label{eq:Newell}
\end{equation}
where $w$ is the speed of the backward propagating congestion waves in traffic.
Newell's model allows one to predict the ego vehicle's future motion, which is illustrated graphically in \autoref{fig:shifted_prediction}(c,d).
Given the data about the lead and ego vehicles' past motion (shown by black and blue curves) up to the time $t$ of prediction (see vertical line), one can identify the time shift $T(t)$ between the trajectories by numerically solving~(\ref{eq:Newell}) for $T(t)$.
Then, the ego's future motion at $t+\theta$, ${\theta \in [0,T(t)]}$ (shown by green curve) is predicted by translating the lead vehicle's past velocity (see red curve) according to:
\begin{equation}\label{eq:Newell_pred_cont}
\begin{split}
\hat{v}_{\rm E}(t,\theta) = \tilde{v}_{\rm E}(t,\theta) & = v_{\rm L}(t+\theta-T(t)),
\end{split}
\end{equation}
where we use tilde $\tilde{v}_{\rm E}(t,\theta)$ to distinguish Newell's prediction from other methods.
\autoref{fig:shifted_prediction}(c,d) illustrate that such a simple prediction (with wave speed ${w=5\,{\rm m/s}}$) can qualitatively capture an upcoming slowdown, although there are quantitative errors in the ego's speed preview.


Practically, predictions are made in discrete time, using data at time steps ${t_i=i \Delta t}$, ${\theta_j=j \Delta t}$, ${i,j \in \mathbb{Z}}$, where the sampling time is ${\Delta t = 0.1\,{\rm s}}$ in our dataset.
For example, when predicting $l$ time steps of future motion, the constant \eqref{eq:const_pred_cont} and Newell \eqref{eq:Newell_pred_cont} predictions in discrete time are
\begin{equation}
\begin{bmatrix}
\hat{v}_{\rm E}(i,l) \\
\vdots \\
\hat{v}_{\rm E}(i,2) \\
\hat{v}_{\rm E}(i,1)
\end{bmatrix}
 =
\begin{bmatrix}
 v_{\rm E}(i) \\
 \vdots \\
 v_{\rm E}(i) \\
 v_{\rm E}(i)
\end{bmatrix}, \quad
\begin{bmatrix}
\tilde{v}_{\rm E}(i,l) \\
\vdots \\
\tilde{v}_{\rm E}(i,2) \\
\tilde{v}_{\rm E}(i,1)
\end{bmatrix}
 =
 \begin{bmatrix}
 v_{\rm L}(i+l-\sigma) \\
 \vdots \\
 v_{\rm L}(i+2-\sigma) \\
 v_{\rm L}(i+1-\sigma)
\end{bmatrix},
\end{equation}
respectively, where $\hat{v}_{\rm E}(i,j)$ is used as a short notation for $\hat{v}_{\rm E}(t_i,\theta_j)$.
Here ${\sigma=T(t)/\Delta t}$ is the maximum achievable horizon, thus ${l \leq \sigma}$ must hold.
We consider ${l=400}$, that is, 40 seconds of prediction horizon in what follows.

Apart from Newell's model predictions, one may directly use deep learning  to achieve more accurate results.
As a baseline, we consider an algorithm which does not use data from V2V connectivity but relies solely on the ego vehicle's data.
This method, called \texttt{Ego-only LSTM-FC}, predicts $l$ time steps of future motion using $k$ time steps of past data according to
\begin{equation}
\begin{bmatrix}
\hat{v}_{\rm E}(i,l) \\
\vdots \\
\hat{v}_{\rm E}(i,2) \\
\hat{v}_{\rm E}(i,1)
\end{bmatrix}
 = M_1 \left(
 \begin{bmatrix}
 v_{\rm E}(i) \\
 v_{\rm E}(i-1) \\
 \vdots \\
 v_{\rm E}(i-k+1)
\end{bmatrix} \right),
\end{equation}
where we used ${k=600}$, ${l=400}$, while $M_1$ is the map underlying the deep learning architecture outlined below.
Note that constant prediction is a special case where $M_1$ outputs values of $v_{\rm E}(i)$.

\paragraph{Hybrid first principle-deep learning  methods}
We propose a hybrid method which uses deep learning to improve the prediction $\tilde{v}_{\rm E}(t,\theta)$ output by Newell's model.
We thus leverage the insights of the first principle-based model and focus on learning higher order fluctuations in the signal.
We explore two configurations: (i) using the Newell's model prediction as an additional input feature and (ii) outputting the residual between the Newell's model prediction and the truth.

In configuration (i), the input
consists of two time series: $k$ time steps of the ego's past velocity
and $l$ time steps of the Newell prediction.
The output
is $l$ time steps of the ego's predicted future velocity.
The input-output relationship is given by a map $M_2$ as summarized by
\begin{equation}
\begin{bmatrix}
\hat{v}_{\rm E}(i,l) \\
\vdots \\
\hat{v}_{\rm E}(i,2) \\
\hat{v}_{\rm E}(i,1)
\end{bmatrix}
 = M_2 \left(
 \begin{bmatrix}
 v_{\rm E}(i) \\
 v_{\rm E}(i-1) \\
 \vdots \\
 v_{\rm E}(i-k+1)
\end{bmatrix},
 \begin{bmatrix}
 \tilde{v}_{\rm E}(i,l) \\
 \vdots \\
 \tilde{v}_{\rm E}(i,2) \\
 \tilde{v}_{\rm E}(i,1)
\end{bmatrix} \right),
\end{equation}
where
we used ${k=600}$, ${l=400}$.
We implemented this configuration on two architectures: a fully-connected feed-forward neural network, called \texttt{Vel-FC}, and an LSTM network, called \texttt{VelLSTM-FC}; see details about the model architecture below. 

For configuration (ii), the input
also contains the residual error
$R(i,j) = v_{\rm E}(i+j) - \tilde{v}_{\rm E}(i,j)$
of Newell's model over the past.
The output
is the next $l$ time steps of the predicted residual
$\hat{R}(i,j)$.
Then $\hat{v}_{\rm E}$ can be reconstructed as
$\hat{v}_{\rm E}(i,j) = \hat{R}(i,j) + \tilde{v}_{\rm E}(i,j)$.
This can be written formally as
\begin{align}
\begin{split}
R(i,j) & =v_{\rm E}(i+j) - \tilde{v}_{E}(i,j),
\quad j \in \{ -\bar{k}, \ldots, 1, 0 \}, \\
\begin{bmatrix}
\hat{R}(i,l) \\
\vdots \\
\hat{R}(i,2) \\
\hat{R}(i,1)
\end{bmatrix}
& = M_3 \left(
 \begin{bmatrix}
 v_{\rm E}(i) \\
 v_{\rm E}(i-1) \\
 \vdots \\
 v_{\rm E}(i-k+1)
\end{bmatrix},
 \begin{bmatrix}
 \tilde{v}_{\rm E}(i,l) \\
 \vdots \\
 \tilde{v}_{\rm E}(i,-\underline{k}+2) \\
 \tilde{v}_{\rm E}(i,-\underline{k}+1)
\end{bmatrix},
 \begin{bmatrix}
 R(i,0) \\
 R(i,-1) \\
 \vdots \\
 R(i,-k+1)
\end{bmatrix} \right), \\
\hat{v}_{\rm E}(i,j) & = \tilde{v}_{\rm E}(i,j) + \hat{R}(i,j),
\quad j \in \{ 1, 2, \ldots, l \},
\end{split}
\end{align}
where
we used ${k=600}$, ${\underline{k}=300}$, ${l=400}$.
Here $M_3$ denotes the map underlying the deep learning architecture, which is an LSTM network called \texttt{ResLSTM-FC}.

\paragraph{Model Architecture}
We implement the functions $M_1,M_2$ and $M_3$ 
using the same encoder-decoder architecture. We use a two-layer LSTM encoder \citep{hochreiter1997long} with \texttt{ReLU} and $\mathrm{tanh}$ activation functions to encode the input into a hidden state vector
$h = \texttt{LSTM-Enc}(\mathbf{x})$
where we initialize the hidden state vector randomly.  In each case the input vectors are concatenated into a single long sequence. For \texttt{VelLSTM-FC},
$\mathbf{x} = [v_E(i-k+1:i)\ \ \tilde{v}_E(i,1:l)]$
and for \texttt{ResLSTM-FC},
$\mathbf{x} = [v_E(i-k+1:i)\ \ \tilde{v}_E(i,-\underline{k}+1:l)\ \ R(i,-{k}+1:0)]$.
This structure accommodates different numbers of time steps for different inputs.    The hidden state vector $h$ is then passed to a linear decoder layer which predicts $l$ future time steps in one shot
$\mathbf{y} = \texttt{Dec}(h)$.


\begin{figure}
    \centering
    \includegraphics[width=130mm]{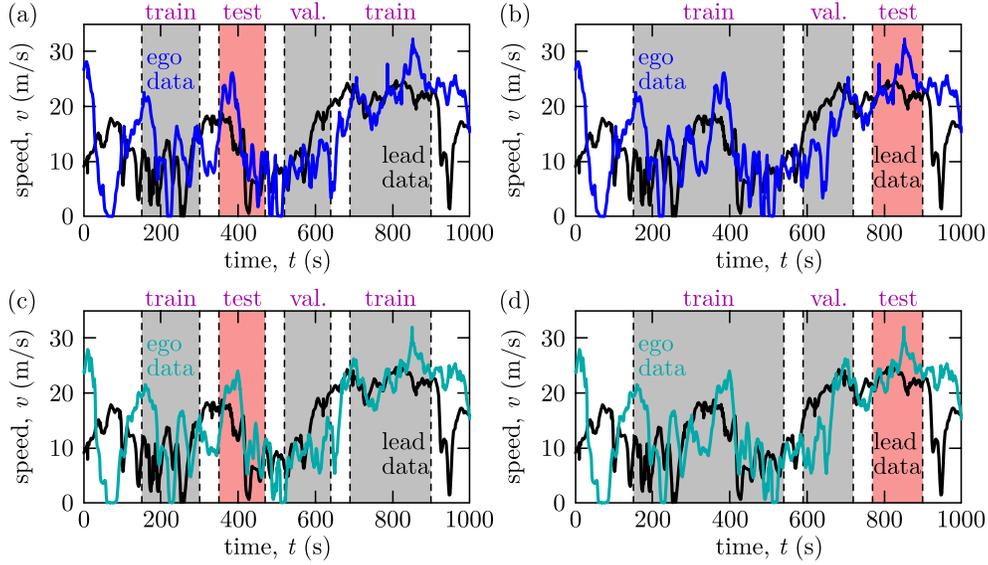}
    \caption{Illustration of the data obtained from two lead-ego pairs (top and bottom) and its split into training, validation and test sets in two alternative ways (left and right).}
    \label{fig:datasets}
    \vspace{-5mm}
\end{figure}




\section{Results}

\paragraph{Data Preparation and Split}
We constructed samples from the data using rolling windows with 600 input time steps (representing 60 seconds of data) and 400 output time steps (40 seconds of data) for two vehicle pairs.
We used 50 second buffers between train, validation, and test sets in order to ensure the test data is not partially
seen during training.  
We created two train-validation-test splits shown in \autoref{fig:datasets}(a,c) and \autoref{fig:datasets}(b,d) corresponding to sample sizes 7200/2400/2400 and 7800/2600/2600, respectively.
Our primary split, shown in panels (a) and (c), has the test set during a traffic jam with slower speed and varying traffic conditions.
These are the conditions under which Newell's model applies and thus the target domain for our method.
To test robustness, we also use an alternative split as shown in panels (b) and (d) where
the test set falls in a regime with higher speed and less speed variation.
The data was normalized by subtracting the sample mean and dividing by the standard deviation of the train set, and we denormalized our results using the inverse formulation.

\begin{figure}[!t]
    \centering
    \includegraphics[width=130mm]{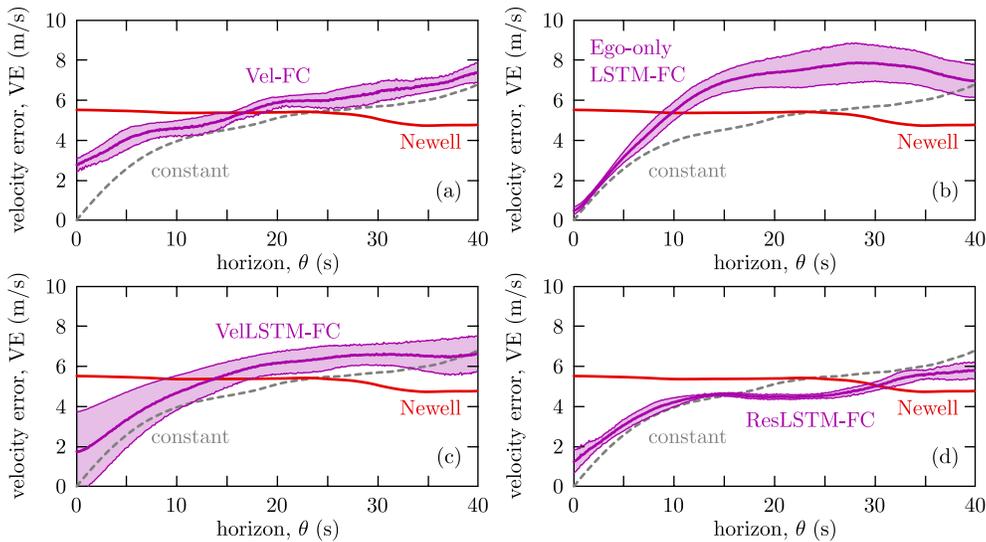}
    \caption{Velocity error ${\rm VE}$ of the (a) \texttt{Vel-FC}, (b) \texttt{Ego-only LSTM-FC}, (c) \texttt{VelLSTM-FC} and (d) \texttt{ResLSTM-FC} models compared to the constant and Newell prediction baselines.
    Results are averaged over two lead-ego pairs and five trained networks using our primary data split.
    The standard deviation amongst the five networks is indicated by shading.}
    \label{fig:velocity_error}
\end{figure}

\paragraph{Evaluation Metrics}  We use the absolute prediction error (PE)
\begin{equation}
{\rm PE}(i,j) = |\hat{v}_{\rm E}(i,j) - v_{\rm E}(i+j)|,
\end{equation}
to compute two performance metrics called velocity error (VE) and average velocity error (AVE):
\begin{equation}
{\rm VE}(j) = \frac{1}{|\mathcal{I}|} \sum_{i \in \mathcal{I}} {\rm PE}(i,j), \quad {\rm AVE} = \frac{1}{l} \sum_{j=1}^{l} {\rm VE}(j) = \frac{1}{l |\mathcal{I}|} \sum_{j=1}^{l} \sum_{i \in \mathcal{I}} {\rm PE}(i,j).
\end{equation}
The velocity error at horizon $j$   is the average of the absolute prediction error amongst the samples in a certain dataset (e.g., training, validation or test set) denoted by $\mathcal{I}$ with $|\mathcal{I}|$ number of elements.
The average velocity error, on the other hand, averages results over the prediction horizon $l$ as well.


\begin{figure}[!t]
    \centering
    \includegraphics[width=135mm]{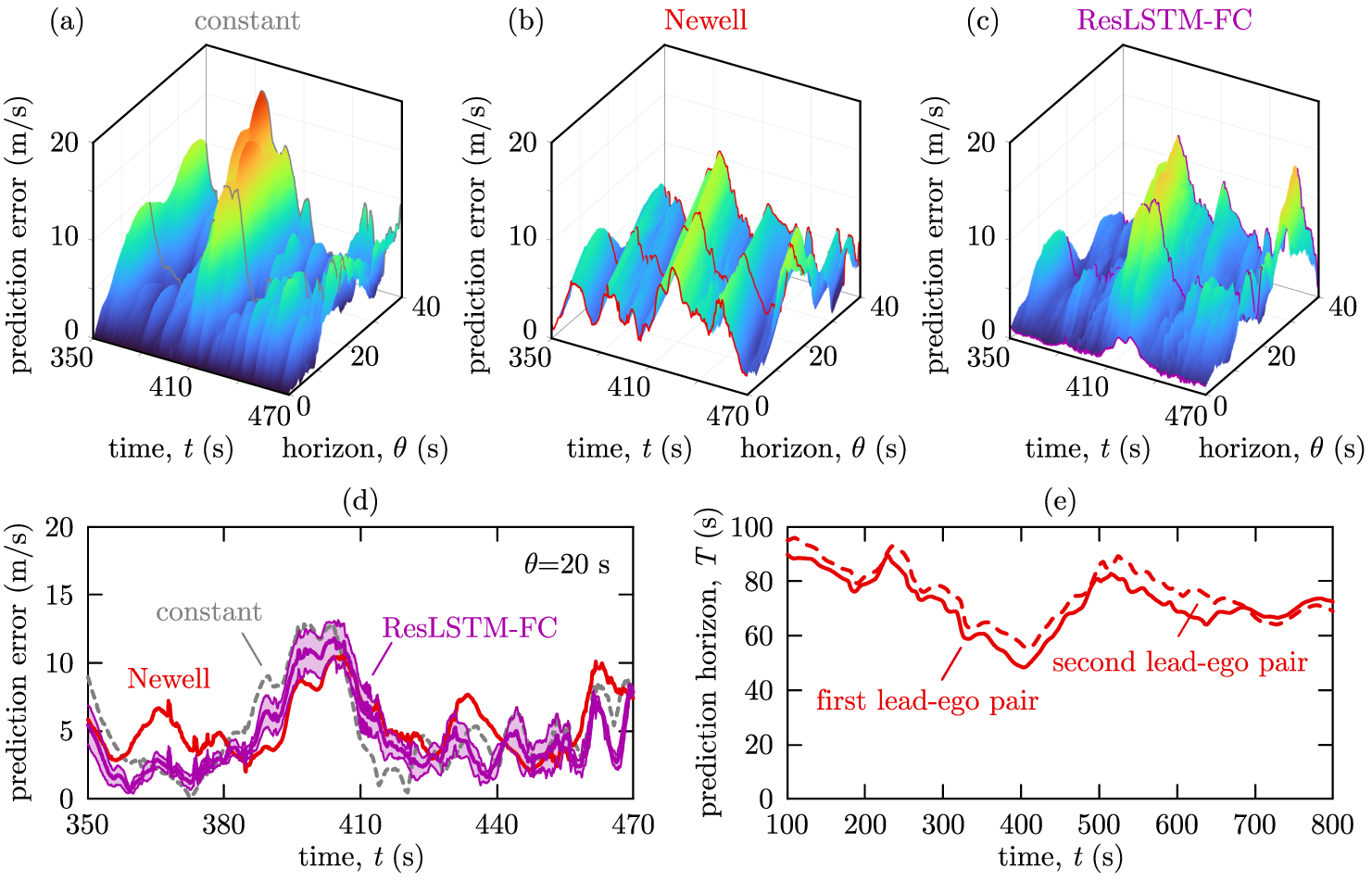}
    \caption{Error metrics evaluated for constant prediction, Newell's model and the \texttt{ResLSTM-FC} model.
    Errors are averaged over two lead-ego pairs.
    Surface of the absolute prediction error for (a) constant (b) Newell's (c) \texttt{ResLSTM-FC} prediction.
    (d) Section of the error surfaces at ${\theta=20\,{\rm s}}$.
    (e) The maximum achievable prediction horizon $T(t)$ based on Newell's model for both lead-ego pairs.}
    \label{fig:ml_vs_baseline}
    \vspace{-5mm}
\end{figure}

\begin{table}
\small
\begin{center}
\begin{tabular}{l|cccc|c}
\toprule
Method & VE@10s & VE@20s & VE@30s & VE@40s & AVE@40s   \\
\midrule
Constant Velocity   & $\mathbf{3.94}$ & $5.10$ & $5.69$ & $6.78$ & $4.61$ \\
Newell (translated)   &  $5.37$ & $5.40$  &  $5.08$ & $\mathbf{4.77}$ & $5.24$ \\
\midrule 
\texttt{Vel-FC} & $4.59\pm0.38$ & $5.88\pm0.27$ & $6.36\pm0.48$ & $7.38\pm0.49$ & $5.48\pm0.15$ \\
\texttt{Ego-only LSTM-FC} & $5.45\pm0.44$ & $7.33\pm0.63$ & $7.86\pm0.86$ & $7.03\pm0.81$ & $6.23\pm0.48$ \\
\texttt{VelLSTM-FC}  & $4.67 \pm 0.86$ & $6.15 \pm 0.54$ & $6.63 \pm 0.54$ & $6.63 \pm 0.88$ & $5.43 \pm 0.75$ \\
\texttt{ResLSTM-FC} & $4.24 \pm 0.31$ & $\mathbf{4.49} \pm 0.18$ & $\mathbf{4.98} \pm 0.26 $ & $5.79 \pm 0.57$ &  $\mathbf{4.34} \pm 0.14$ \\
\bottomrule
\end{tabular}
\caption{Results for main test set from $t=350$ to $t=470$ seconds as shown in \autoref{fig:datasets}(a,c). All numbers are in m/s. Forecasting accuracy at $\theta=10,20,30,40$ seconds and mean accuracy over 40 seconds is shown.  Errors are shown as mean plus minus one standard deviation over five training runs with random seeds.}
\label{table:results_early_test}
\vspace{-10mm}
\end{center}
\end{table}

\paragraph{Prediction Accuracy}
The results for our primary data split are detailed in \autoref{table:results_early_test} and visualized in \autoref{fig:velocity_error}.
The velocity error decreases over the prediction horizon with different rates for different methods.
The constant velocity metric gives the smallest velocity error for short horizon, but the error grows quickly.
Conversely, the error for the Newell prediction starts somewhat higher but remains constant over time.
These baselines are outperformed by our best performing model \texttt{RelLSTM-FC}, whose error is close to or better than the baselines across the forecasting window.
In particular, it blends the positive characteristics of both baselines: having small error for small $\theta$ and low error growth for higher $\theta$. 
\texttt{RelLSTM-FC} has slightly higher error than constant prediction for small $\theta$, it has lower error than both baselines over $15 \leq \theta \leq 30$, while it is surpassed by Newell's model for larger $\theta$.
Note that it is not possible to predict beyond $\theta > T(t)$ since at this point the traffic conditions of the ego vehicle have not yet been encountered by the lead vehicle. 

Meanwhile, the \texttt{Ego-only LSTM-FC} architecture, which takes only the ego car's velocity as input, has worse performance than both first principle-based methods and \texttt{RelLSTM-FC}.
This justifies that using data from V2V connectivity has significant benefits and also validates our hybrid approach.  
Lastly, we compare to a monolithic fully connected network \texttt{Vel-FC} which has the same input and output as \texttt{VelLSTM-FC} (see Appendix \ref{app:vel-fc}).
These networks have similar performance, and they are both outperformed by  \texttt{ResLSTM-FC} that gives the overall best average velocity error.
These suggest that the most significant parts of our method are integrating deep learning with the first principle-based Newell prediction and predicting the {\em residual} from Newell's model as output. 

\begin{figure}[t!]
    \centering
    \includegraphics[width=130mm]{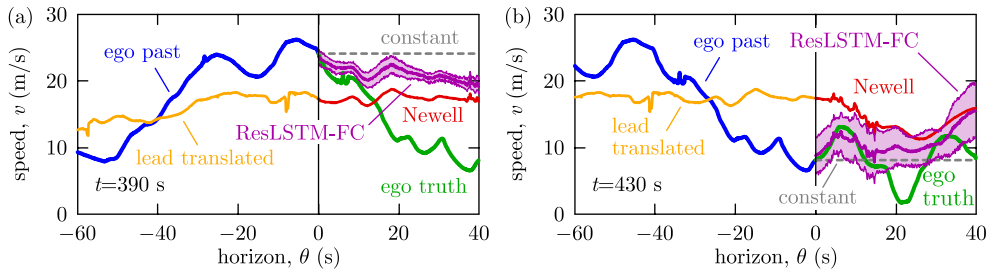}
    \caption{Illustration of the predictions provided by constant velocity, Newell and \texttt{ResLSTM-FC} models for two selected samples at (a) ${t = 390\,{\rm s}}$ and (b) ${t = 430\,{\rm s}}$ for one lead-ego pair.}
    \label{fig:samples}
\end{figure}

\autoref{fig:ml_vs_baseline}(a,b,c) provide further visualization of the prediction accuracy by showing the absolute prediction error for the constant prediction, Newell's model and \texttt{ResLSTM-FC}.
\autoref{fig:ml_vs_baseline}(d) shows their comparison for a selected horizon, while \autoref{fig:ml_vs_baseline}(e) indicates the physically achievable maximum horizon $T(t)$.
Sample predictions are visualized in \autoref{fig:samples}.
For each network architecture, training was repeated five times.
In all figures purple line shows the mean and shading indicates plus minus one standard deviation over five trained networks.



\paragraph{Accuracy under Distributional Shift} To investigate the robustness of our method, we also test using our alternative data split where training and validation data are in the middle of the traffic jam characterized by variable and slower speeds.  However, the test set is from a period of faster and more free flowing traffic. 
Results are shown in \autoref{table:results_late_test}.  Velocity errors are smaller across all methods for this test set since cars have little variability in speed.   The performance of the deep learning methods is worse relative to constant velocity and Newell baselines since small variations in the speed are hard to predict.  However, we do see that our methods generalize reasonably well over the distributional shift as their overall error goes down despite the higher magnitude of the velocity outputs.


\begin{table}[t!]
\small
\begin{center}
\begin{tabular}{l|cccc|c}
\toprule
Method & VE@10s & VE@20s & VE@30s & VE@40s & AVE@40s   \\
\midrule
Constant Velocity   & $\mathbf{1.85}$ & $\mathbf{2.51}$ & $\mathbf{2.83}$ & $3.31$ & $\mathbf{2.25}$ \\
Newell (translated)   &  $2.86$ & $2.98$  & $3.08$  & $\mathbf{2.93}$ & $2.96$ \\
\midrule 
\texttt{Vel-FC} & $2.65 \pm 0.10$ & $3.81 \pm 0.14$ & $4.51 \pm 0.22$ & $4.45 \pm 0.51$ & $3.49 \pm 0.14$ \\
\texttt{Ego-only LSTM-FC} & $2.18 \pm 0.27$ & $3.05 \pm 0.24$ & $3.67 \pm 0.18$ & $4.45 \pm 0.29$ & $2.97 \pm 0.28$ \\
\texttt{VelLSTM-FC} & $2.74 \pm 0.96$ & $3.45 \pm 1.18$ & $3.97 \pm 0.89$ & $4.88 \pm 0.85$ & $3.25 \pm 0.77$ \\
\texttt{ResLSTM-FC} & $2.19 \pm 0.29$ & $2.62 \pm 0.18$
 & $3.28 \pm 0.2$ & $4.16 \pm 0.48$ & $2.61 \pm 0.16$   \\
\bottomrule
\end{tabular}
\caption{Results for alternative test set
from ${t=770}$ to ${t=900}$ seconds.
As shown by \autoref{fig:datasets}(b,d), there is a distributional shift between training and testing data which is especially challenging to predict.  All numbers are in m/s.  Forecasting accuracy at $\theta=10,20,30,40$ seconds and mean accuracy over 40 seconds is shown.  Errors are shown as mean plus minus one standard deviation over five training runs with random seeds.}
\label{table:results_late_test}
\end{center}
\vspace{-10mm}
\end{table}

\section{Discussion}

We integrated first principle models and data-driven deep learning  for traffic prediction utilizing vehicle-to-vehicle communication. The proposed model (\texttt{ResLSTM-FC}) showed improved accuracy over both purely first principle-based and purely deep learning-based baselines. This model used a prediction from Newell's model as input to an LSTM neural network and the prediction of the residual error as output. 
Unlike baselines, \texttt{ResLSTM-FC} has both a small error for short-term predictions and a slowly growing error as the forecasting horizon increases.
The model was trained and tested on raw GPS data, and
the observed robustness makes it a feasible candidate for 
real-time on-board traffic predictions for connected vehicles.

Our model achieved fairly good generalization under distributional shift, however, first principle baselines generalize better as they are independent of the data distribution.
We hypothesize that further improvements in our model can be obtained by more careful corrections of errors in the underlying first principle model.   
Namely, while the Newell's model prediction often succeeds in predicting the degree of a slowdown, it is often slightly early or late.  This corresponds to a time-shift error.  Future work includes adding prediction of this time shift to \texttt{ResLSTM-FC}. 
Another potential area for improvement would be to replace the Newell's model with a more sophisticated first principle baseline, and to consider data from multiple lead vehicles for prediction.

\acks{
This research was partially supported by the University of Michigan's Center of Connected and Automated Transportation through the US DOT grant 69A3551747105, Google Faculty Research Award, NSF Grant \#2037745,
and the U. S. Army Research Office under Grant W911NF-20-1-0334.
}

\bibliography{ref.bib}

\appendix

\section{Hyperparameter Tuning}
We tuned hyperparameters over the ranges shown in \autoref{tab:hyper}. 
\begin{table}[ht]
    \centering
    \begin{tabular}{ll}
    \toprule
        Hyperparameter & Range \\
        \midrule
        Input Time Steps ($k$) &  300, 400, 500, \textbf{600} \\
         LSTM Hidden Units & 10, 20, 50, 100, \textbf{200} \\
        Learning Rate & \textbf{0.0005}, 0.001, 0.005, 0.01   \\        
        Prediction Horizon ($l$) & 50, 100, 200, 300, \textbf{400} \\
        \bottomrule
    \end{tabular}
    \caption{Hyperparameter tuning ranges.  Bold values correspond to best performance.}
    \label{tab:hyper}
\end{table}

\section{Vel-FC Model Architecture}
\label{app:vel-fc}
The model architecture of \texttt{Vel-FC} is shown in \autoref{tab:FC_arc}. 
\begin{table}[ht]
    \centering
    \begin{tabular}{ll}
    \toprule
        Layer & Hyperparameters \\
        \midrule
        Linear 1 & in\_features=1000, out\_features=200, bias=True, followed by ReLU \\
        Linear 2 & in\_features=200, out\_features=200, bias=True, followed by ReLU \\
        Linear 3 & in\_features=200, out\_features=400, bias=True \\        
        \bottomrule
    \end{tabular}
    \caption{Model architecture of \texttt{Vel-FC}.}
    \label{tab:FC_arc}
\end{table}
\end{document}